\PassOptionsToPackage{table,xcdraw}{xcolor}
\documentclass[sigconf,nonacm]{acmart}

\usepackage{amsmath}
\usepackage{amsthm}
\usepackage{booktabs}
\usepackage{algorithm}
\usepackage{algpseudocode}
\usepackage{array}
\usepackage{multirow}
\usepackage{tabularx}
\usepackage{threeparttable}
\usepackage{siunitx}
\usepackage{lscape}
\usepackage{soul}
\usepackage{newfloat}
\usepackage{listings}
\usepackage{marvosym}

\usepackage[normalem]{ulem}

\useunder{\uline}{\ul}{}

\graphicspath{{Figures/}}
\setcopyright{none}

\renewcommand\footnotetextcopyrightpermission[1]{}
\acmConference[]{}{}{}

\title{\texorpdfstring{S$^3$-Diff}{S3-Diff}: Structural Semantic Synergy Diffusion Model for High Fidelity Super Resolution of Pathological Images}

\author{Jiaming Liang\textsuperscript{\textdagger}}
\orcid{0009-0000-6007-9759}
\affiliation{\institution{South China University of Technology}\department{School of Computer Science and Engineering}\city{Guangzhou}\country{China}}

\author{QiHui Han\textsuperscript{\textdagger}}
\orcid{0009-0007-2727-1889}
\affiliation{\institution{South China University of Technology}\department{Software Engineering}\city{Guangzhou}\country{China}}

\author{Guangye Ou}
\orcid{0009-0009-5541-2372}
\affiliation{\institution{South China University of Technology}\department{School of Computer Science}\city{Guangzhou}\country{China}}

\author{Jiawen Liu}
\orcid{0009-0002-6827-6049}
\affiliation{\institution{South China University of Technology}\department{School of Computer Science and Engineering}\city{Guangzhou}\country{China}}

\author{Haolin Chen}
\orcid{0009-0001-1831-5276}
\affiliation{\institution{South China University of Technology}\department{Future Technology School}\city{Guangzhou}\country{China}}

\author{Xi Zhong}
\orcid{0000-0002-0713-105X}
\affiliation{\institution{Affiliated Cancer Hospital of Guangzhou Medical University}\department{Department of Radiology}\city{Guangzhou}\country{China}}

\author{Jiazhou Chen\textsuperscript{\Letter}}
\orcid{0000-0001-7171-9547}
\affiliation{\institution{Guangdong University of Technology}\department{School of Computer Science and Technology}\city{Guangzhou}\country{China}}

\email{csjzchen@gdut.edu.cn}

\author{Xiaoqi Sheng\textsuperscript{\Letter}}
\orcid{0000-0002-2929-5805}
\affiliation{\institution{South China University of Technology}\department{School of Future Technology}\city{Guangzhou}\country{China}}

\email{xqsheng@scut.edu.cn}
\author{Hongmin Cai}
\orcid{0000-0002-2747-7234}
\affiliation{\institution{South China University of Technology}\department{School of Future Technology}\city{Guangzhou}\country{China}}

\begin{document}

\begin{abstract}
Digital pathology relies on high-resolution whole slide images for accurate diagnosis, yet limitations in imaging devices, storage, and transmission often make lower-resolution pathology images more common in clinical workflows. Current super-resolution techniques often tend to smooth diagnostically relevant morphology, leading to over-smoothed textures and semantic drift that compromise downstream clinical interpretation. To this end, we develop the Structural Semantic Synergy Diffusion Model (S$^3$-Diff), a diffusion framework for high-fidelity super-resolution of pathological images. The core of S$^3$-Diff is Specimen-aware Structural Anchoring (SSA), which combines prognosis-aware tissue support extracted by a fixed SAM with LR-HR gradient discrepancies to generate a specimen-specific structural anchor to preserve pathological morphology. Concurrently, we introduce Structure-guided Semantic Fidelity Tuning (SSFT) to adapt DINOv3 representations using SSA-derived structural supervision. SSFT combines the adapted semantic energy with LR-derived edge and grayscale cues. The resulting control guides denoising to suppress stochastic artifacts and maintain structural consistency. Extensive experimental results demonstrate that S$^3$-Diff consistently outperforms state-of-the-art methods in both reconstruction quality and downstream survival analysis performance. The source code will be made public.
\end{abstract}


\maketitle
\vspace{0pt}
\noindent
\textsuperscript{\textdagger} Equal contribution.
\quad
\textsuperscript{\Letter} Corresponding authors.
\section{Introduction}
Precision medicine relies on accurate disease characterization for diagnosis, prognosis, and individualized treatment~\cite{wang2023precision}. In this process, Digital Pathology (DP) has become an important data foundation for precision medicine. Whole Slide Images (WSIs), which provide high-resolution representations of tissue morphology, are widely regarded as a gold standard for diagnosis as they preserve rich morphological information for clinical evaluation~\cite{bejnordi2017diagnostic}. However, the clinical deployment of High-Resolution (HR) WSIs remains constrained by substantial storage costs, demanding data-transmission bandwidth, and the limited availability of high-magnification scanners in resource-constrained settings~\cite{zarella2019practical}. Consequently, Low-Resolution (LR) WSIs are often adopted in practical clinical workflows~\cite{zehra2023suggested}. Nevertheless, LR images may lose pathological structures required for manual diagnosis and downstream computer-aided analysis~\cite{shafi2023artificial}, motivating cost-effective methods to recover diagnostically relevant details without additional imaging hardware.

\begin{figure}[!t]
    \centering
    \includegraphics[width=1.0\linewidth]{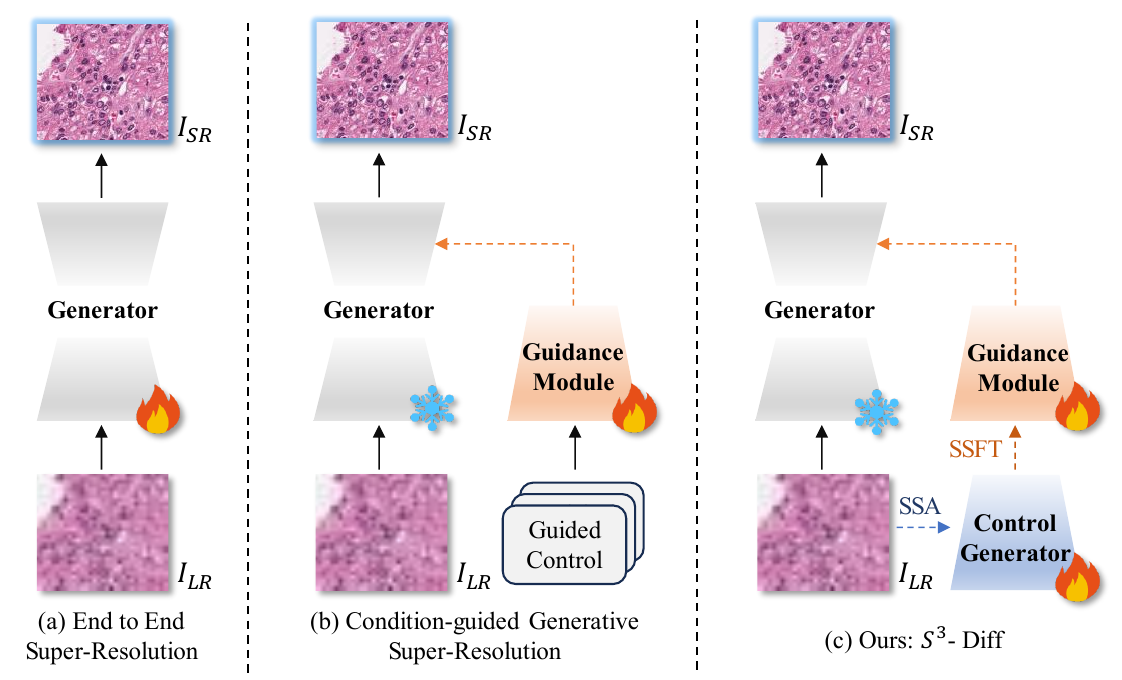}
    \caption{Comparison of architectural designs between traditional pathological image super-resolution frameworks and our proposed S$^3$-Diff.}
    \label{fig: fig1}
\Description{Diagram comparing a conventional pathology super-resolution pipeline with the proposed structure- and semantics-guided S3-Diff pipeline.}
\end{figure}


Super-resolution (SR) provides a cost-effective solution by reconstructing fine pathological details from LR observations~\cite{lepcha2023image}. The ability to enhance routinely accessible LR scans makes SR particularly valuable for extending high-quality pathological assessment to settings with limited imaging resources. Existing SR methods can be broadly divided into two categories~\cite{su2025review}. As illustrated in Fig.~\ref{fig: fig1}(a), the first category learns deterministic LR-to-HR mappings by minimizing pixel-level reconstruction losses. Representative methods include SwinIR~\cite{liang2021swinir}, SHISRCNet~\cite{xie2023shisrcnet}, CWT-Net~\cite{jia2024cwt}, and MiHATP~\cite{xu2024mihatp}, which employ Transformer-based regression or pathology-aware structural priors to restore fine cellular and stromal details. However, reconstruction losses do not explicitly distinguish diagnostically salient morphology from surrounding tissue and may average plausible high-frequency patterns, producing over-smoothed textures with diminished diagnostic utility~\cite{jia2024cwt}.


As shown in Fig.~\ref{fig: fig1}(b), the second category comprises generative SR methods based on GANs or diffusion models, including ESRGAN~\cite{wang2018esrgan}, SinSR~\cite{wang2024sinsr}, and SuperDiff~\cite{xu2025superdiff}. By synthesizing high-frequency textures under conditional guidance, generative approaches can substantially improve perceptual realism~\cite{liang2022details}. However, stochastic generation, training-distribution dependence, and weak geometric constraints may undermine structural consistency and alter diagnostically relevant morpholog~\cite{zhang2025pixel}. Thus, pathology SR thus faces a fundamental mismatch between visually plausible detail synthesis and morphology-faithful reconstruction.


Accordingly, we propose the Structural Semantic Synergy Diffusion Model (S$^3$-Diff), illustrated in Fig.~\ref{fig: fig1}(c). The framework combines training-time geometric supervision with trainable LR-derived semantic control to promote structural fidelity and semantic consistency. Specifically, Specimen-aware Structural Anchoring (SSA) uses a fixed Segment Anything Model (SAM)~\cite{kirillov2023segment} and clinical risk priors to construct prognosis-aware structural reference maps that emphasize morphology attenuated by LR degradation. Complementarily, DINOv3 supplies robust dense semantics, but its generic representations lack explicit sensitivity to pathology-specific geometry. Structure-guided Semantic Fidelity Tuning (SSFT) therefore uses SSA-derived geometric supervision to adapt these representations toward tissue morphology and combines the resulting semantic energy with edge and grayscale cues to construct the composite control flow. At inference, this LR-derived control guides latent denoising without requiring SAM or HR targets.

Evaluations on The Cancer Genome Atlas (TCGA~\cite{weinstein2013cancer}) datasets show that S$^3$-Diff achieves strong structural and perceptual fidelity, improving LPIPS and ST-LPIPS by up to 23.49\% and 23.63\% over the strongest competing methods. On the independent SurGen cohort, excluded from training and checkpoint selection, S$^3$-Diff maintains competitive perceptual quality and achieves the lowest Grad-L1 among learned methods, suggesting cross-cohort robustness under the evaluated protocol. Downstream survival analysis further demonstrates that S$^3$-Diff preserves clinically relevant semantic information in the reconstructed SR images. Our contributions are summarized as follows:

\begin{itemize}
    \item \textbf{Structure-Semantic Diffusion Framework}: 
    We introduce S$^3$-Diff, a structure-semantic synergy diffusion framework for pathology image SR that couples generative reconstruction with geometric and semantic constraints to recover faithful pathological details.
    
    \item \textbf{Collaborative Structural-Semantic Guidance}: 
    We introduce SSA and SSFT to propagate SAM-derived structural supervision into DINOv3 semantic representations, providing LR-informed control for artifact-suppressed latent denoising and morphology-preserving reconstruction.
    \item \textbf{Evaluation and Downstream Validation}: 
    Experiments demonstrate that S$^3$-Diff delivers superior structural preservation and perceptual fidelity compared with existing SR methods. Downstream evaluation further confirms that the reconstructed images preserve survival-relevant information under the adopted validation protocol.
\end{itemize}

\section{Related Work}

\subsection{Deterministic Pathology SR and Structural Detail Loss}

Deterministic SR reconstructs an HR image from its LR counterpart using end-to-end encoder-decoders optimized for pixel-wise fidelity~\cite{wang2019sr_survey}. Building on convolutional architectures~\cite{dong2015image}, attention-based models~\cite{li2023multi}, including HAT~\cite{chen2025hat} and SwinIR~\cite{liang2021swinir}, capture long-range dependencies to strengthen feature representation.

For pathology images, SHISRCNet~\cite{xie2023shisrcnet}, MiHATP~\cite{xu2024mihatp}, and CWT-Net~\cite{jia2024cwt} introduce multi-scale fusion or wavelet transforms to capture complex tissue structure. These priors improve tissue-aware reconstruction without departing from a deterministic regression objective. Although effective on pixel-level metrics, their regression-to-the-mean objectives tend to smooth high-frequency textures and fine cellular details that support diagnosis. Moreover, one-to-one mappings inadequately represent the structural uncertainty of pathological tissue~\cite{su2025review,xu2025superdiff}, motivating models that preserve both local morphology and global structure.

\begin{figure*}[!t]
    \centering      
    \includegraphics[width=\textwidth]{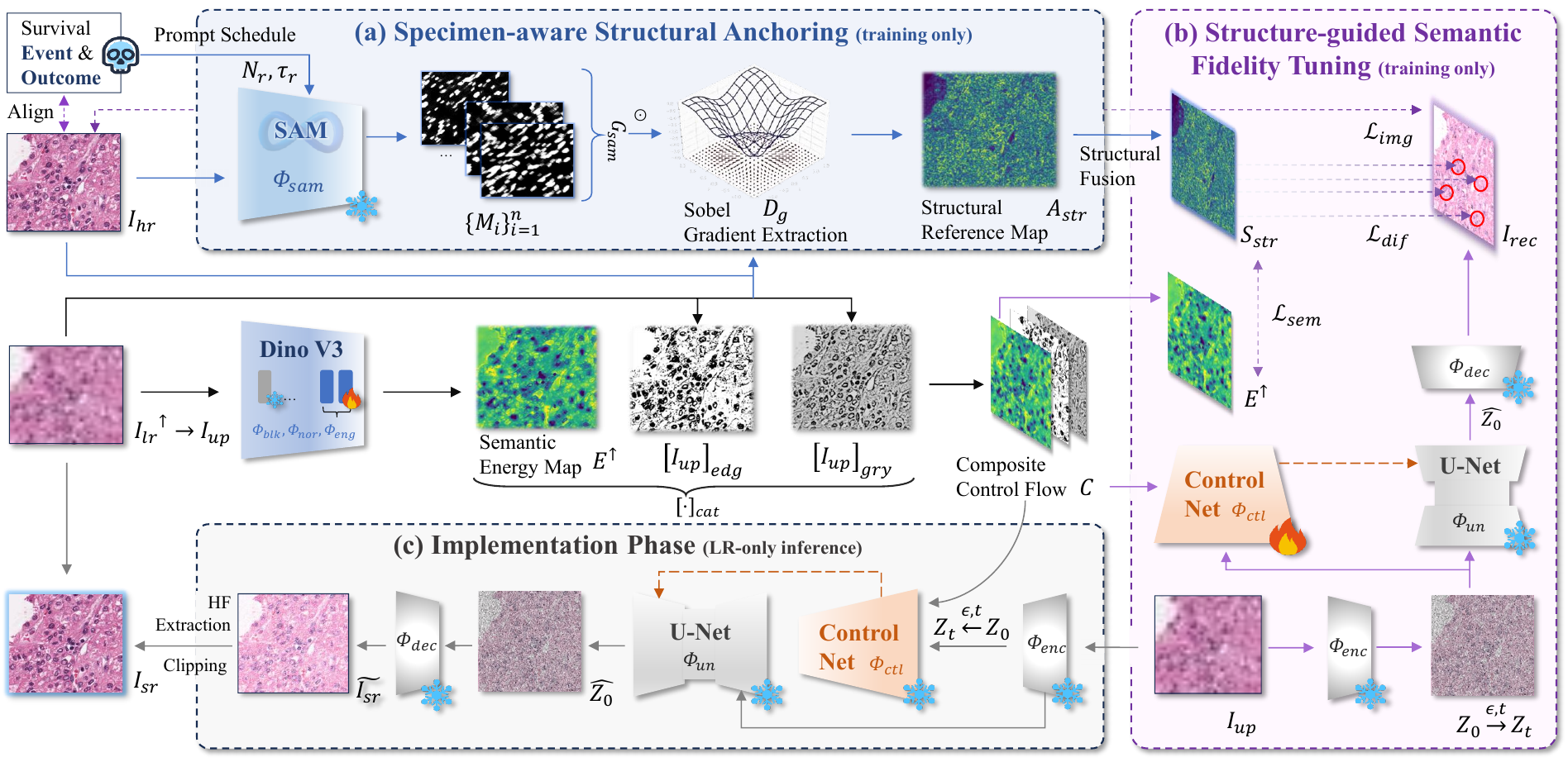}
    \caption{Framework Overview of S$^3$-Diff.}
    \label{fig: overview}
\Description{Block diagram of the S3-Diff architecture, including specimen-aware structural anchoring and structure-guided semantic fidelity tuning.}
\end{figure*}

\subsection{Generative Pathology SR and Structural Fidelity}

Generative SR uses GANs and diffusion models to synthesize fine textures and improve perceptual quality~\cite{wang2021real,yang2023diffusion}. ESRGAN~\cite{wang2018esrgan} demonstrates stronger detail generation than reconstruction-based methods, addressing the oversmoothing of deterministic regression. However, transferring such models to pathology is risky because statistically plausible textures may not faithfully reproduce biological structures~\cite{croitoru2023diffusion}, making structural reliability dependent on explicit constraints.

Recent diffusion-based SR methods improve perceptual realism through iterative denoising, but their stochastic generation process often lacks sufficient pathology-aware structural constraints. Constrained or domain-specific approaches, including SinSR~\cite{wang2024sinsr}, SuperDiff~\cite{xu2025superdiff}, and URCDM~\cite{cechnicka2024urcdm}, mitigate stochastic hallucinations and structural distortions through priors or sampling optimization. However, reconciling global topological preservation with high-frequency texture recovery remains nontrivial~\cite{su2025review}. Perceptually convincing outputs may still distort disease-relevant tissue organization, underscoring the need for semantic control grounded in pathological morphology~\cite{han2026beyond}. Meanwhile, the extent to which structural fidelity supports downstream clinical inference, including survival prediction, remains unclear. An effective pathology SR framework should consequently align perceptual reconstruction, structural integrity, and clinically meaningful semantics.

\section{Methodology}

\subsection{Preliminaries}

Pathology image SR learns a mapping from an LR input $I_{lr}\in\mathbb{R}^{3\times H\times W}$ to an SR output $I_{sr}\in\mathbb{R}^{3\times4H\times4W}$ under the supervision of its HR counterpart $I_{hr}\in\mathbb{R}^{3\times4H\times4W}$. The HR image $I_{hr}$ is available only during training, while inference reconstructs $I_{sr}$ from $I_{lr}$ alone. S$^3$-Diff adopts latent diffusion~\cite{rombach2022high} to balance reconstruction quality and computational cost. A pre-trained VAE encoder $\Phi_{enc}(\cdot)$ maps $I_{hr}$ into the clean latent representation $Z_0=\Phi_{enc}(I_{hr})$. The diffusion process then operates in this compact latent space. At timestep $t$, Gaussian noise $\epsilon\sim\mathcal{N}(0,\mathbf{I})$ is injected according to:
\begin{equation}
Z_t
=
\sqrt{\bar{\alpha}_t}Z_0
+
\sqrt{1-\bar{\alpha}_t}\epsilon,
\label{eq:forward_diffusion}
\end{equation}
where $\bar{\alpha}_t$ denotes the cumulative coefficient determined by the noise schedule. The matrix $\mathbf{I}$ denotes the identity covariance of the Gaussian distribution. During reverse denoising, the trainable ControlNet $\Phi_{ctl}(\cdot)$ maps $Z_t$, $t$, and composite
control flow $C$ to residuals for the frozen U-Net $\Phi_{un}(\cdot)$~\cite{zhang2023controlnet}. The U-Net predicts the injected noise $\widehat{\epsilon}_{C}$. The control flow $C$ combines LR-derived structure with pathology-adapted semantics. SSFT evaluates this prediction against $\epsilon$ and updates the control and semantic components, while the U-Net remains frozen.

\subsection{Overview of the Framework}
The proposed Structural Semantic Synergy Diffusion Model (S$^3$-Diff) coordinates structural and semantic constraints throughout latent denoising, yielding morphology-faithful and semantically consistent reconstructions. The framework is explicitly designed for the information asymmetry between training and inference, where paired HR images provide rich structural supervision during optimization, whereas deployment relies solely on $I_{lr}$. As shown in Figure~\ref{fig: overview}, $I_{lr}$ is first upsampled into the bicubic reference $I_{up}$, which provides the LR reference for both stages. During training, SSA uses prognosis-aware prompting with a fixed SAM $\Phi_{sam}(\cdot)$~\cite{kirillov2023segment} to extract tissue support from $I_{hr}$. SSA combines this support with the gradient discrepancy between $I_{hr}$ and $I_{up}$ to produce $A_{str}$. Used only during training, $A_{str}$ emphasizes structural details attenuated by downsampling and supervises semantic adaptation and denoising optimization. Under this structural supervision, SSFT constructs $C$ entirely from $I_{up}$. The LR-derived control flow guides latent denoising and remains available at inference without HR information.

The LR-derived control flow $C$ carries the learned structural and semantic guidance into latent reconstruction. The trainable ControlNet $\Phi_{ctl}(\cdot)$~\cite{zhang2023controlnet} maps $C$ into multi-scale residuals that are injected into the frozen U-Net $\Phi_{un}(\cdot)$. During training, the forward process in Equation~\eqref{eq:forward_diffusion} perturbs the HR latent $Z_0$ into $Z_t$. Optimization under $A_{str}$ supervision aligns the HR latent target with the LR-derived control. At inference, $\Phi_{enc}(I_{up})$ is perturbed with strength $\xi_i$, which defaults to $0.20$. The same $C$ is reused throughout reverse diffusion. The VAE decoder $\Phi_{dec}(\cdot)$ reconstructs the preliminary image $\widetilde{I}_{sr}$. A two-stage LR-anchored reconstruction combines its diffusion-restored detail with the LR-derived context in $I_{up}$ to produce $I_{sr}$. Thus, HR-derived structural priors guide learning without requiring $A_{str}$, $I_{hr}$, or prognostic metadata at inference.

\subsection{Specimen-Aware Structural Anchoring}

As shown in Figure~\ref{fig: overview}(a), SSA constructs a specimen-aware structural anchor from HR morphology under prognostic guidance. This anchor constrains latent denoising to reduce structural shifts. Since the edge discrepancy between $I_{hr}$ and $I_{up}$ may also respond to irrelevant background texture, a fixed SAM $\Phi_{sam}(\cdot)$~\cite{kirillov2023segment} is introduced to extract tissue regions and contour support from $I_{hr}$. Its zero-shot segmentation capability provides stable tissue support for complex pathological morphology~\cite{ma2024segment}. Given that different prognostic outcomes are associated with heterogeneous nuclear and immune morphology~\cite{lauss2024molecular,song2023artificial}, SSA incorporates prognostic risk into specimen-aware structural anchoring. The coefficient $R\in[0,1]$ is derived from survival time $T_s$ and death-event status $\delta_s$:
\begin{equation}
R=
\begin{cases}
(1+\gamma_rT_s)^{-1}, & \delta_s=1,\\
0, & \delta_s=0,
\end{cases}
\label{eq:risk}
\end{equation}
where $T_s$ denotes survival time in days, $\delta_s=1$ indicates an observed death event, and $\delta_s=0$ denotes censoring. The coefficient $\gamma_r$ controls sensitivity to survival time and defaults to $0.01$. For observed events, shorter survival yields a larger $R$.

The risk coefficient adapts the spatial coverage of SAM support used for structural anchoring. The foreground prompt count $N_r$ controls sampling density, while $\tau_r$ filters masks by confidence. Under the default linear schedule, $N_r$ increases from $4$ to $32$ as $R$ increases and is rounded to the nearest integer. Over the same range, $\tau_r$ decreases from $0.90$ to $0.55$. The resulting prognosis-aware masks are defined as:
\begin{equation}
\{M_i\}_{i=1}^{n}
=
\Phi_{sam}(I_{hr},N_r,\tau_r),
\label{eq:sam_masks}
\end{equation}
where $n$ denotes the number of retained masks. This schedule makes the SAM tissue support responsive to prognosis-associated structural heterogeneity. The complete process remains confined to offline preprocessing.

The retained masks are merged and expanded to form $G_{sam}$, which defines the valid tissue region for structural anchoring. The discrepancy map $D_g$ localizes structural detail present in $I_{hr}$ but attenuated in $I_{up}$. Their spatial overlap forms the final structural anchor $A_{str}$:
\begin{equation}
\begin{gathered}
G_{sam}
=
\left[\displaystyle\cup_{i=1}^{n}M_i\right]_{dil},
\\[2pt]
D_g
=
\left[\left\lvert
[I_{hr}]_{sob}-[I_{up}]_{sob}
\right\rvert\right]_{mm},
\\[2pt]
A_{str}
=
\left[G_{sam}\odot D_g\right]_{mm}.
\end{gathered}
\label{eq:ssa_anchor}
\end{equation}
Here, $\cup_{i=1}^{n}M_i$ denotes the union of the retained SAM masks. The operator $[\cdot]_{dil}$ denotes morphological dilation with a default radius of three pixels. The operator $[\cdot]_{sob}$ extracts Sobel gradients from the channel-mean grayscale image~\cite{zhao2024msef}. The operator $[\cdot]_{mm}$ denotes per-image min-max normalization, while $\odot$ denotes element-wise multiplication. The resulting $A_{str}$ retains downsampling-induced structural discrepancies within prognosis-adapted tissue support. It anchors semantic adaptation and diffusion optimization to structurally relevant pathological regions.

\subsection{Structure-Guided Semantic Fidelity Tuning}

Since HR-derived structural supervision is unavailable at inference, SSFT transfers this training-only structural information from SSA into semantic control derived from the LR input. As shown in Figure~\ref{fig: overview}(b), $I_{up}$ is spatially aligned with $I_{hr}$ and resized before being encoded by the VFM encoder $\Phi_{vfm}(\cdot)$, which is instantiated with a pre-trained DINOv3 ViT-B/16~\cite{simeoni2025dinov3}. The Energy Head $\Phi_{eng}(\cdot)$ maps the resulting patch features into the normalized semantic energy map $E$. To preserve pre-trained visual priors, adaptation updates $\Phi_{eng}$, the final Transformer block $\Phi_{blk}$, and all LayerNorm modules $\Phi_{nor}$, while keeping the remaining backbone frozen. SSFT combines the upsampled energy map with edge and grayscale cues from $I_{up}$ to construct:
\begin{equation}
C=
\left[
E^{\uparrow},
[I_{up}]_{edg},
[I_{up}]_{gry}
\right]_{cat},
\label{eq:composite_control}
\end{equation}
where $E^{\uparrow}$ denotes bilinear upsampling of $E$ to the spatial dimensions of $I_{up}$. The operators $[\cdot]_{edg}$, $[\cdot]_{gry}$, and $[\cdot]_{cat}$ denote finite-difference edge extraction, channel-mean grayscale projection, and channel-wise concatenation, respectively. At each timestep, the trainable ControlNet $\Phi_{ctl}(\cdot)$ receives $Z_t$, $t$, and $C$ to produce multi-scale residuals. The frozen U-Net $\Phi_{un}(\cdot)$ processes the same $Z_t$ and $t$ together with these residuals to predict $\widehat{\epsilon}_{C}$. The LR-derived $C$ is computed once for each input and reused throughout reverse diffusion.

The LR-derived control flow $C$ remains available during inference. In contrast, $A_{str}$ provides training-only structural supervision. SSFT transfers this structural information into semantic adaptation by combining $A_{str}$ with the HR and UP edge maps:
\begin{equation}
S_{str}
=
\left[
\rho_sA_{str}
+\rho_h[I_{hr}]_{edg}
+\rho_u[I_{up}]_{edg}
\right]_{[0,1]},
\label{eq:structure_target}
\end{equation}
where $[\cdot]_{[0,1]}$ denotes clipping to the unit interval. The coefficients $\rho_s$, $\rho_h$, and $\rho_u$ weight the structural reference, HR edges, and UP edges, respectively, with default values of $0.60$, $0.30$, and $0.10$. The resulting $S_{str}$ encodes the spatial importance of morphological boundaries that require faithful recovery. During optimization, $S_{str}$ provides spatial weights for the pixel and edge terms in $\mathcal{L}_{img}$. It also supervises $E^{\uparrow}$. The structural anchor $A_{str}$ separately reweights $\mathcal{L}_{dif}$ at latent resolution.

To expose controlled denoising to image-domain supervision, SSFT converts the conditional noise prediction into a clean latent and a training reconstruction:
\begin{equation}
\begin{gathered}
\widehat{Z}_0
=
\frac{
Z_t-\sqrt{1-\bar{\alpha}_t}\widehat{\epsilon}_{C}
}{
\sqrt{\bar{\alpha}_t}
},
\\
I_{rec}
=
\Phi_{dec}(\widehat{Z}_0).
\end{gathered}
\label{eq:training_reconstruction}
\end{equation}
SSFT uses $I_{rec}$ only during training rather than as the final inference output $I_{sr}$. The decoder $\Phi_{dec}(\cdot)$ remains frozen but provides a differentiable path through which image-domain errors reach the trainable control and semantic branches.

SSFT jointly optimizes denoising accuracy, reconstruction fidelity, and semantic alignment:
\begin{equation}
\begin{gathered}
\mathcal{L}_{dif}
=
\mathbb{E}_{Z_0,\epsilon,t}
\left\langle
[A_{str}]_{lat}\odot
(\epsilon-\widehat{\epsilon}_{C})^2
\right\rangle,
\\
\mathcal{L}_{img}
=
[I_{rec},I_{hr},S_{str}]_{fid},
\\
\mathcal{L}_{sem}
=
\left\langle
\left|E^{\uparrow}-S_{str}\right|
\right\rangle,
\\
\mathcal{L}_{tot}
=
\mathcal{L}_{dif}
+\lambda_i\mathcal{L}_{img}
+\lambda_s\mathcal{L}_{sem}.
\end{gathered}
\label{eq:ssft_total_loss}
\end{equation}
Here, $\mathbb{E}_{Z_0,\epsilon,t}$ denotes expectation over the clean HR latent, sampled Gaussian noise, and diffusion timestep, while $\langle\cdot\rangle$ denotes element-wise averaging. The operation $[A_{str}]_{lat}$ projects $A_{str}$ to latent resolution and applies scheduled mean normalization. Intuitively, $\mathcal{L}_{dif}$ emphasizes accurate denoising at SSA-selected structures. The image-fidelity term $\mathcal{L}_{img}$ uses $[\cdot,\cdot,\cdot]_{fid}$ to combine $S_{str}$-weighted pixel and edge errors with low-frequency color and LPIPS discrepancies~\cite{zhang2018perceptual}, thereby preserving morphology, staining, and perceptual texture. The semantic term $\mathcal{L}_{sem}$ aligns $E^{\uparrow}$ with $S_{str}$, converting SSA geometry into pathology-aware semantic attention. The coefficients $\lambda_i$ and $\lambda_s$ weight image fidelity and semantic consistency, with default values of $0.12$ and $0.04$. Gradients from $\mathcal{L}_{dif}$ and $\mathcal{L}_{img}$ reach ControlNet and the adapted DINOv3 branch, whereas $\mathcal{L}_{sem}$ directly tunes only the semantic branch. This complementary optimization embeds SSA geometry into both controlled denoising and LR-derived semantic guidance.

\subsection{Training and Implementation}

During training, paired $I_{lr}$ and $I_{hr}$ provide the conditional input and reconstruction target. Bicubic upsampling converts $I_{lr}$ into $I_{up}$, while $\Phi_{enc}(\cdot)$ maps $I_{hr}$ to $Z_0$. Randomly sampled $\epsilon$ and $t$ then produce $Z_t$ according to Equation~\eqref{eq:forward_diffusion}, and $I_{up}$ generates $C$ through Equation~\eqref{eq:composite_control}. This construction aligns LR-derived conditioning with the HR latent target. The pre-trained latent-diffusion backbone remains frozen to preserve its generative prior, so optimization is confined to $\Phi_{ctl}(\cdot)$ and the trainable SSFT components described above. These components are jointly optimized by $\mathcal{L}_{tot}$ in Equation~\eqref{eq:ssft_total_loss}.

After training, all model parameters are frozen, and inference requires only $I_{lr}$. Bicubic upsampling first produces $I_{up}$, from which the LR-derived control flow $C$ is constructed. In parallel, the frozen VAE encoder $\Phi_{enc}(\cdot)$ maps $I_{up}$ into a reference latent, which is perturbed with strength $\xi_i$ to initialize reverse diffusion. The same $C$ guides every reverse step without requiring $A_{str}$, $\Phi_{sam}(\cdot)$, $I_{hr}$, or prognostic metadata. The denoised latent is then decoded by $\Phi_{dec}(\cdot)$ into the preliminary image $\widetilde{I}_{sr}$. This image provides diffusion-restored morphological detail, whereas $I_{up}$ preserves the LR-derived low-frequency staining and spatial context. Their complementary information is integrated through the following two-stage LR-anchored reconstruction:
\begin{equation}
\begin{gathered}
I_{sr}^{(1)}
=
\left[
I_{up}
+
[\widetilde{I}_{sr}]_{hf}^{(1)}
\right]_{[0,1]},
\\
I_{sr}
=
\left[
I_{up}
+
[I_{sr}^{(1)}]_{hf}^{(2)}
\right]_{[0,1]}.
\end{gathered}
\label{eq:cascade_refinement}
\end{equation}
Here, $[\cdot]_{hf}^{(1)}$ denotes first-stage high-frequency extraction, while $[\cdot]_{hf}^{(2)}$ denotes bounded second-stage refinement with hard clipping. The detailed perturbation, filtering, gain, and clipping settings are provided in Appendix Section A. The first stage transfers diffusion-derived high-frequency information to $I_{up}$, and the bounded second stage limits excessive sharpening before final LR anchoring. This design carries the learned structural-semantic guidance into inference using only $I_{lr}$.

\section{Experiments}

\begin{table*}[!t]
\centering
\renewcommand{\arraystretch}{0.95}
\caption{The SR performance of various frameworks utilizing the TCGA and SurGen dataset. The best and second-best results are highlighted in \textbf{bold} and \underline{underlined}, respectively, while our proposed method is shaded in gray. Meanwhile, {\color[HTML]{CB0000}\ensuremath{\uparrow}} represents the percentage performance improvement of S$^3$-Diff compared to each comparison method, while {\color[HTML]{036400}\ensuremath{\downarrow}} represents the percentage decrease.}
\resizebox{0.92\textwidth}{!}{
\begin{tabular}{l|lrrrrr}
\toprule[0.8mm]
\multirow{2}{*}{\textbf{Datasets}} & \multirow{2}{*}{\textbf{Methods}} & \multicolumn{5}{c}{\textbf{Metrics}} \\ \cline{3-7}
 & & LPIPS\ensuremath{\downarrow} & ST-LPIPS\ensuremath{\downarrow} & Grad-L1\ensuremath{\downarrow} & PSNR\ensuremath{\uparrow} & SSIM\ensuremath{\uparrow} \\
\midrule[0.5mm]
\multirow{9}{*}{\rotatebox{90}{\textbf{TCGA-LUAD}}} & Bicubic & 0.3423 {\color[HTML]{CB0000}\ensuremath{\uparrow}51.77\%} & 0.3624 {\color[HTML]{CB0000}\ensuremath{\uparrow}54.61\%} & 0.0326 {\color[HTML]{CB0000}\ensuremath{\uparrow}77.61\%} & \underline{26.23}  {\color[HTML]{CB0000}\ensuremath{\uparrow}11.25\%} & \underline{0.7067} {\color[HTML]{CB0000}\ensuremath{\uparrow}16.02\%} \\
 & SHISRCNet~\cite{xie2023shisrcnet} & 0.2504 {\color[HTML]{CB0000}\ensuremath{\uparrow}34.07\%} & 0.2525 {\color[HTML]{CB0000}\ensuremath{\uparrow}34.85\%} & 0.0232 {\color[HTML]{CB0000}\ensuremath{\uparrow}68.53\%} & 22.62 {\color[HTML]{CB0000}\ensuremath{\uparrow}29.00\%} & 0.5646 {\color[HTML]{CB0000}\ensuremath{\uparrow}45.22\%} \\
 & SwinIR~\cite{liang2021swinir} & 0.2424 {\color[HTML]{CB0000}\ensuremath{\uparrow}31.89\%} & 0.2466 {\color[HTML]{CB0000}\ensuremath{\uparrow}33.29\%} & 0.0212 {\color[HTML]{CB0000}\ensuremath{\uparrow}65.57\%} & 22.42 {\color[HTML]{CB0000}\ensuremath{\uparrow}30.15\%} & 0.5827 {\color[HTML]{CB0000}\ensuremath{\uparrow}40.71\%} \\
 & ESRGAN~\cite{wang2018esrgan} & 0.2666 {\color[HTML]{CB0000}\ensuremath{\uparrow}38.07\%} & 0.2666 {\color[HTML]{CB0000}\ensuremath{\uparrow}38.30\%} & 0.0184 {\color[HTML]{CB0000}\ensuremath{\uparrow}60.33\%} & 23.66 {\color[HTML]{CB0000}\ensuremath{\uparrow}23.33\%} & 0.6426 {\color[HTML]{CB0000}\ensuremath{\uparrow}27.59\%} \\
 & SinSR~\cite{wang2024sinsr} & 0.2444 {\color[HTML]{CB0000}\ensuremath{\uparrow}32.45\%} & 0.2452 {\color[HTML]{CB0000}\ensuremath{\uparrow}32.91\%} & \underline{0.0146} {\color[HTML]{CB0000}\ensuremath{\uparrow}50.00\%} & 25.44 {\color[HTML]{CB0000}\ensuremath{\uparrow}14.70\%} & 0.6817 {\color[HTML]{CB0000}\ensuremath{\uparrow}20.27\%} \\
 & STAR-RL~\cite{chen2024star} & \underline{0.2158} {\color[HTML]{CB0000}\ensuremath{\uparrow}23.49\%} & \underline{0.2154} {\color[HTML]{CB0000}\ensuremath{\uparrow}23.63\%} & 0.0199 {\color[HTML]{CB0000}\ensuremath{\uparrow}63.32\%} & 24.21 {\color[HTML]{CB0000}\ensuremath{\uparrow}20.53\%} & 0.6784 {\color[HTML]{CB0000}\ensuremath{\uparrow}20.86\%} \\
 & UPSR~\cite{zhang2025uncertainty} & 0.3011 {\color[HTML]{CB0000}\ensuremath{\uparrow}45.17\%} & 0.3121 {\color[HTML]{CB0000}\ensuremath{\uparrow}47.29\%} & 0.0252 {\color[HTML]{CB0000}\ensuremath{\uparrow}71.03\%} & 18.65 {\color[HTML]{CB0000}\ensuremath{\uparrow}56.46\%} & 0.5454 {\color[HTML]{CB0000}\ensuremath{\uparrow}50.33\%} \\
 & SuperDiff~\cite{xu2025superdiff} & 0.2624 {\color[HTML]{CB0000}\ensuremath{\uparrow}37.08\%} & 0.2532 {\color[HTML]{CB0000}\ensuremath{\uparrow}35.03\%} & 0.0188 {\color[HTML]{CB0000}\ensuremath{\uparrow}61.17\%} & 23.73 {\color[HTML]{CB0000}\ensuremath{\uparrow}22.97\%} & 0.6527 {\color[HTML]{CB0000}\ensuremath{\uparrow}25.62\%} \\
 & \cellcolor[HTML]{EFEFEF}S$^3$-Diff (ours) & \cellcolor[HTML]{EFEFEF}\textbf{0.1651} & \cellcolor[HTML]{EFEFEF}\textbf{0.1645} & \cellcolor[HTML]{EFEFEF}\textbf{0.0073} & \cellcolor[HTML]{EFEFEF}\textbf{29.18} & \cellcolor[HTML]{EFEFEF}\textbf{0.8199} \\
\midrule[0.5mm]
\multirow{9}{*}{\rotatebox{90}{\textbf{TCGA-KIRC}}} & Bicubic & 0.3521 {\color[HTML]{CB0000}\ensuremath{\uparrow}43.71\%} & 0.3386 {\color[HTML]{CB0000}\ensuremath{\uparrow}41.73\%} & 0.1020 {\color[HTML]{CB0000}\ensuremath{\uparrow}88.04\%} & \underline{26.23} {\color[HTML]{CB0000}\ensuremath{\uparrow}11.06\%} & \underline{0.7021} {\color[HTML]{CB0000}\ensuremath{\uparrow}13.12\%} \\
 & SHISRCNet~\cite{xie2023shisrcnet} & 0.2504 {\color[HTML]{CB0000}\ensuremath{\uparrow}20.85\%} & 0.2525 {\color[HTML]{CB0000}\ensuremath{\uparrow}21.86\%} & 0.0232 {\color[HTML]{CB0000}\ensuremath{\uparrow}47.41\%} & 23.04 {\color[HTML]{CB0000}\ensuremath{\uparrow}26.43\%} & 0.5645 {\color[HTML]{CB0000}\ensuremath{\uparrow}40.69\%} \\
 & SwinIR~\cite{liang2021swinir} & 0.2204 {\color[HTML]{CB0000}\ensuremath{\uparrow}10.07\%} & 0.2211 {\color[HTML]{CB0000}\ensuremath{\uparrow}10.76\%} & 0.0222 {\color[HTML]{CB0000}\ensuremath{\uparrow}45.05\%} & 24.06 {\color[HTML]{CB0000}\ensuremath{\uparrow}21.07\%} & 0.6941 {\color[HTML]{CB0000}\ensuremath{\uparrow}14.42\%} \\
 & ESRGAN~\cite{wang2018esrgan} & 0.2121 {\color[HTML]{CB0000}\ensuremath{\uparrow}\phantom{0}6.55\%} & 0.2121 {\color[HTML]{CB0000}\ensuremath{\uparrow}\phantom{0}6.98\%} & 0.0161 {\color[HTML]{CB0000}\ensuremath{\uparrow}24.22\%} & 22.86 {\color[HTML]{CB0000}\ensuremath{\uparrow}27.43\%} & 0.6253 {\color[HTML]{CB0000}\ensuremath{\uparrow}27.01\%} \\
 & SinSR~\cite{wang2024sinsr} & 0.2552 {\color[HTML]{CB0000}\ensuremath{\uparrow}22.34\%} & 0.2564 {\color[HTML]{CB0000}\ensuremath{\uparrow}23.05\%} & 0.0154 {\color[HTML]{CB0000}\ensuremath{\uparrow}20.78\%} & 25.15 {\color[HTML]{CB0000}\ensuremath{\uparrow}15.83\%} & 0.6614 {\color[HTML]{CB0000}\ensuremath{\uparrow}20.08\%} \\
 & STAR-RL~\cite{chen2024star} & 0.2358 {\color[HTML]{CB0000}\ensuremath{\uparrow}15.95\%} & 0.2361 {\color[HTML]{CB0000}\ensuremath{\uparrow}16.43\%} & 0.0234 {\color[HTML]{CB0000}\ensuremath{\uparrow}47.86\%} & 26.12 {\color[HTML]{CB0000}\ensuremath{\uparrow}11.52\%} & 0.6823 {\color[HTML]{CB0000}\ensuremath{\uparrow}16.40\%} \\
 & UPSR~\cite{zhang2025uncertainty} & 0.2848 {\color[HTML]{CB0000}\ensuremath{\uparrow}30.41\%} & 0.2848 {\color[HTML]{CB0000}\ensuremath{\uparrow}30.72\%} & 0.0242 {\color[HTML]{CB0000}\ensuremath{\uparrow}49.59\%} & 19.55 {\color[HTML]{CB0000}\ensuremath{\uparrow}49.00\%} & 0.5555 {\color[HTML]{CB0000}\ensuremath{\uparrow}42.97\%} \\
 & SuperDiff~\cite{xu2025superdiff} & \underline{0.2104} {\color[HTML]{CB0000}\ensuremath{\uparrow}\phantom{0}5.80\%} & \underline{0.2104} {\color[HTML]{CB0000}\ensuremath{\uparrow}\phantom{0}6.23\%} & \textbf{0.0120} {\color[HTML]{036400}\ensuremath{\downarrow}\phantom{0}1.67\%} & 23.62 {\color[HTML]{CB0000}\ensuremath{\uparrow}23.33\%} & 0.6747 {\color[HTML]{CB0000}\ensuremath{\uparrow}17.71\%} \\
 & \cellcolor[HTML]{EFEFEF}S$^3$-Diff (ours) & \cellcolor[HTML]{EFEFEF}\textbf{0.1982} & \cellcolor[HTML]{EFEFEF}\textbf{0.1973} & \cellcolor[HTML]{EFEFEF}\underline{0.0122} & \cellcolor[HTML]{EFEFEF}\textbf{29.13} & \cellcolor[HTML]{EFEFEF}\textbf{0.7942} \\
\midrule[0.5mm]
\multirow{9}{*}{\rotatebox{90}{\textbf{TCGA-LIHC}}} & Bicubic & 0.3523 {\color[HTML]{CB0000}\ensuremath{\uparrow}47.43\%} & 0.3384 {\color[HTML]{CB0000}\ensuremath{\uparrow}45.45\%} & 0.1020 {\color[HTML]{CB0000}\ensuremath{\uparrow}91.08\%} & \underline{26.23} {\color[HTML]{CB0000}\ensuremath{\uparrow}\phantom{0}8.73\%} & \underline{0.7023} {\color[HTML]{CB0000}\ensuremath{\uparrow}12.83\%} \\
 & SHISRCNet~\cite{xie2023shisrcnet} & 0.2321 {\color[HTML]{CB0000}\ensuremath{\uparrow}20.21\%} & 0.2251 {\color[HTML]{CB0000}\ensuremath{\uparrow}17.99\%} & 0.0182 {\color[HTML]{CB0000}\ensuremath{\uparrow}50.00\%} & 23.24 {\color[HTML]{CB0000}\ensuremath{\uparrow}22.72\%} & 0.6542 {\color[HTML]{CB0000}\ensuremath{\uparrow}21.13\%} \\
 & SwinIR~\cite{liang2021swinir} & \underline{0.2202} {\color[HTML]{CB0000}\ensuremath{\uparrow}15.89\%} & \underline{0.2202} {\color[HTML]{CB0000}\ensuremath{\uparrow}16.17\%} & 0.0221 {\color[HTML]{CB0000}\ensuremath{\uparrow}58.82\%} & 23.12 {\color[HTML]{CB0000}\ensuremath{\uparrow}23.36\%} & 0.6646 {\color[HTML]{CB0000}\ensuremath{\uparrow}19.23\%} \\
 & ESRGAN~\cite{wang2018esrgan} & 0.2332 {\color[HTML]{CB0000}\ensuremath{\uparrow}20.58\%} & 0.2332 {\color[HTML]{CB0000}\ensuremath{\uparrow}20.84\%} & 0.0154 {\color[HTML]{CB0000}\ensuremath{\uparrow}40.91\%} & 23.12 {\color[HTML]{CB0000}\ensuremath{\uparrow}23.36\%} & 0.6324 {\color[HTML]{CB0000}\ensuremath{\uparrow}25.30\%} \\
 & SinSR~\cite{wang2024sinsr} & 0.2652 {\color[HTML]{CB0000}\ensuremath{\uparrow}30.17\%} & 0.2664 {\color[HTML]{CB0000}\ensuremath{\uparrow}30.71\%} & 0.0153 {\color[HTML]{CB0000}\ensuremath{\uparrow}40.52\%} & 25.13 {\color[HTML]{CB0000}\ensuremath{\uparrow}13.49\%} & 0.6514 {\color[HTML]{CB0000}\ensuremath{\uparrow}21.65\%} \\
 & STAR-RL~\cite{chen2024star} & 0.2531 {\color[HTML]{CB0000}\ensuremath{\uparrow}26.83\%} & 0.2531 {\color[HTML]{CB0000}\ensuremath{\uparrow}27.06\%} & 0.0213 {\color[HTML]{CB0000}\ensuremath{\uparrow}57.28\%} & 24.24 {\color[HTML]{CB0000}\ensuremath{\uparrow}17.66\%} & 0.6628 {\color[HTML]{CB0000}\ensuremath{\uparrow}19.55\%} \\
 & UPSR~\cite{zhang2025uncertainty} & 0.2788 {\color[HTML]{CB0000}\ensuremath{\uparrow}33.57\%} & 0.2727 {\color[HTML]{CB0000}\ensuremath{\uparrow}32.31\%} & 0.0224 {\color[HTML]{CB0000}\ensuremath{\uparrow}59.38\%} & 19.34 {\color[HTML]{CB0000}\ensuremath{\uparrow}47.47\%} & 0.5155 {\color[HTML]{CB0000}\ensuremath{\uparrow}53.71\%} \\
 & SuperDiff~\cite{xu2025superdiff} & 0.2492 {\color[HTML]{CB0000}\ensuremath{\uparrow}25.68\%} & 0.2492 {\color[HTML]{CB0000}\ensuremath{\uparrow}25.92\%} & \underline{0.0098} {\color[HTML]{CB0000}\ensuremath{\uparrow}\phantom{0}7.14\%} & 23.62 {\color[HTML]{CB0000}\ensuremath{\uparrow}20.75\%} & 0.6548 {\color[HTML]{CB0000}\ensuremath{\uparrow}21.01\%} \\
 & \cellcolor[HTML]{EFEFEF}S$^3$-Diff (ours) & \cellcolor[HTML]{EFEFEF}\textbf{0.1852} & \cellcolor[HTML]{EFEFEF}\textbf{0.1846} & \cellcolor[HTML]{EFEFEF}\textbf{0.0091} & \cellcolor[HTML]{EFEFEF}\textbf{28.52} & \cellcolor[HTML]{EFEFEF}\textbf{0.7924} \\
\midrule[0.5mm]
\multirow{9}{*}{\rotatebox{90}{\textbf{SurGen}}} & Bicubic & 0.2425 {\color[HTML]{CB0000}\ensuremath{\uparrow}28.04\%} & 0.2392 {\color[HTML]{CB0000}\ensuremath{\uparrow}28.47\%} & 0.0097 {\color[HTML]{CB0000}\ensuremath{\uparrow}67.01\%} & \textbf{38.61} {\color[HTML]{036400}\ensuremath{\downarrow}\phantom{0}9.74\%} & \textbf{0.9248} {\color[HTML]{036400}\ensuremath{\downarrow}\phantom{0}6.57\%} \\
 & SHISRCNet~\cite{xie2023shisrcnet} & 0.2116 {\color[HTML]{CB0000}\ensuremath{\uparrow}17.53\%} & 0.2185 {\color[HTML]{CB0000}\ensuremath{\uparrow}21.69\%} & 0.0082 {\color[HTML]{CB0000}\ensuremath{\uparrow}60.98\%} & 27.51 {\color[HTML]{CB0000}\ensuremath{\uparrow}26.68\%} & 0.6925 {\color[HTML]{CB0000}\ensuremath{\uparrow}24.77\%} \\
 & SwinIR~\cite{liang2021swinir} & 0.1764 {\color[HTML]{CB0000}\ensuremath{\uparrow}\phantom{0}1.08\%} & 0.1781  {\color[HTML]{CB0000}\ensuremath{\uparrow}\phantom{0}3.93\%} & 0.0076 {\color[HTML]{CB0000}\ensuremath{\uparrow}57.89\%} & 30.15 {\color[HTML]{CB0000}\ensuremath{\uparrow}15.59\%} & 0.8224 {\color[HTML]{CB0000}\ensuremath{\uparrow}\phantom{0}5.06\%} \\
 & ESRGAN~\cite{wang2018esrgan} & 0.1784 {\color[HTML]{CB0000}\ensuremath{\uparrow}\phantom{0}2.19\%} & 0.1776 {\color[HTML]{CB0000}\ensuremath{\uparrow}\phantom{0}3.66\%} & 0.0062 {\color[HTML]{CB0000}\ensuremath{\uparrow}48.39\%} & 29.17 {\color[HTML]{CB0000}\ensuremath{\uparrow}19.47\%} & 0.7366 {\color[HTML]{CB0000}\ensuremath{\uparrow}17.30\%} \\
 & SinSR~\cite{wang2024sinsr} & 0.2314 {\color[HTML]{CB0000}\ensuremath{\uparrow}24.59\%} & 0.2324 {\color[HTML]{CB0000}\ensuremath{\uparrow}26.38\%} & 0.0094 {\color[HTML]{CB0000}\ensuremath{\uparrow}65.96\%} & 25.27 {\color[HTML]{CB0000}\ensuremath{\uparrow}37.91\%} & 0.5586 {\color[HTML]{CB0000}\ensuremath{\uparrow}54.67\%} \\
 & STAR-RL~\cite{chen2024star} & 0.1842 {\color[HTML]{CB0000}\ensuremath{\uparrow}\phantom{0}5.27\%} & 0.1842 {\color[HTML]{CB0000}\ensuremath{\uparrow}\phantom{0}7.11\%} & 0.0075 {\color[HTML]{CB0000}\ensuremath{\uparrow}57.33\%} & 30.24 {\color[HTML]{CB0000}\ensuremath{\uparrow}15.24\%} & 0.7566 {\color[HTML]{CB0000}\ensuremath{\uparrow}14.20\%} \\
 & UPSR~\cite{zhang2025uncertainty} & \textbf{0.1723} {\color[HTML]{036400}\ensuremath{\downarrow}\phantom{0}1.28\%} & \underline{0.1735} {\color[HTML]{CB0000}\ensuremath{\uparrow}\phantom{0}1.38\%} & 0.0064 {\color[HTML]{CB0000}\ensuremath{\uparrow}50.00\%} & 28.62 {\color[HTML]{CB0000}\ensuremath{\uparrow}21.77\%} & 0.8125 {\color[HTML]{CB0000}\ensuremath{\uparrow}\phantom{0}6.34\%} \\
 & SuperDiff~\cite{xu2025superdiff} & 0.1782 {\color[HTML]{CB0000}\ensuremath{\uparrow}\phantom{0}2.08\%} & 0.1784 {\color[HTML]{CB0000}\ensuremath{\uparrow}\phantom{0}4.09\%} & \underline{0.0054} {\color[HTML]{CB0000}\ensuremath{\uparrow}40.74\%} & 26.58 {\color[HTML]{CB0000}\ensuremath{\uparrow}31.11\%} & 0.7818 {\color[HTML]{CB0000}\ensuremath{\uparrow}10.51\%} \\
 & \cellcolor[HTML]{EFEFEF}S$^3$-Diff (ours) & \cellcolor[HTML]{EFEFEF}\underline{0.1745} & \cellcolor[HTML]{EFEFEF}\textbf{0.1711} & \cellcolor[HTML]{EFEFEF}\textbf{0.0032} & \cellcolor[HTML]{EFEFEF}\underline{34.85} & \cellcolor[HTML]{EFEFEF}\underline{0.8640} \\
\bottomrule[0.8mm]
\end{tabular}
}
\label{tab:pre}
\end{table*}

\subsection{Experiment Settings}

\textbf{Datasets and Model Configuration}:

We use 150 TCGA WSIs, with 50 from each of the lung (LUAD), kidney (KIRC), and liver (LIHC) cancer cohorts~\cite{weinstein2013cancer}, split at the patient level into training, validation, and test sets at 7:1:2. We additionally use 46 patients from the SurGen colorectal cancer cohort~\cite{myles2025surgen} exclusively for external testing. Cohort selection, partitioning, and preprocessing details are provided in Appendix Section B.1, B.2, and B.3.

S$^3$-Diff uses a frozen Stable Diffusion v1.5 backbone~\cite{rombach2022high} with DINOv3 ViT-B/16~\cite{simeoni2025dinov3} and SAM-derived structural reference maps. Training pairs are constructed by bicubically downsampling $512\times512$ HR patches into $128\times128$ LR inputs, while the model reconstructs $512\times512$ SR outputs. Optimization updates ControlNet, the Energy Head, the final DINOv3 Transformer block, and all DINOv3 LayerNorm parameters. AdamW is applied for 60,000 micro-batch iterations using two-step gradient accumulation, an effective batch size of two, FP16 precision, and seed 2025. ControlNet and the Energy Head use a learning rate of $3\times10^{-6}$, whereas the adapted DINOv3 parameters use $6\times10^{-7}$. The reported checkpoint is selected by the lowest LPIPS on the case-level TCGA validation split, without accessing the TCGA test split or the SurGen cohort. Further training and inference details are provided in Appendix Section A.1 and A.4.

\noindent
\textbf{Evaluation Metrics}:

This work adopts Peak Signal-to-Noise Ratio (PSNR), Structural Similarity Index (SSIM)~\cite{wang2004image}, Learned Perceptual Image Patch Similarity (LPIPS)~\cite{zhang2018perceptual} and its shift-tolerant variant ST-LPIPS, together with the gradient-based structural metric Grad-L1~\cite{xue2014gradient}, for quantitative evaluation. In addition, downstream survival analysis is conducted to assess the prognostic utility of the SR images. For each SR method, an independent attention-based multiple-instance learning model, ABMIL~\cite{ITW:2018}, is trained on its corresponding SR images using an identical case-level data split. The outcome labels used to construct SSA supervision are restricted to the SR training split. No validation or test outcome is used by the SR model during training or inference. The Concordance Index (CI)~\cite{wang2019machine} is adopted to evaluate the ranking accuracy of predicted survival risks. Kaplan-Meier curves~\cite{kaplan1958nonparametric}, constructed by stratifying patients into low-risk and high-risk groups, are further used to qualitatively demonstrate the discriminative ability of the survival prediction results. For detailed information on all indicators, please refer to  Appendix Section C.

\noindent
\textbf{Competing Methods}:

To comprehensively evaluate the performance of the proposed S$^3$-Diff framework, this study compares it with a representative set of competitive SR baselines. According to their modeling paradigms, the compared methods are broadly grouped into two categories: deterministic end-to-end reconstruction methods and generative frameworks. The first category, including SwinIR~\cite{liang2021swinir} and SHISRCNet~\cite{xie2023shisrcnet}, aims to maximize pixel-level structural fidelity through deterministic mappings. The second category, including ESRGAN~\cite{wang2018esrgan}, SinSR~\cite{wang2024sinsr}, STAR-RL~\cite{chen2024star}, UPSR~\cite{zhang2025uncertainty}, and SuperDiff~\cite{xu2025superdiff}, incorporates generative priors for SR reconstruction. Standard Bicubic interpolation is also included as a non-learning baseline by directly upsampling LR images.

\subsection{Results and Discussion}
\begin{figure*}[!t]
    \centering      
    \includegraphics[width=\textwidth]{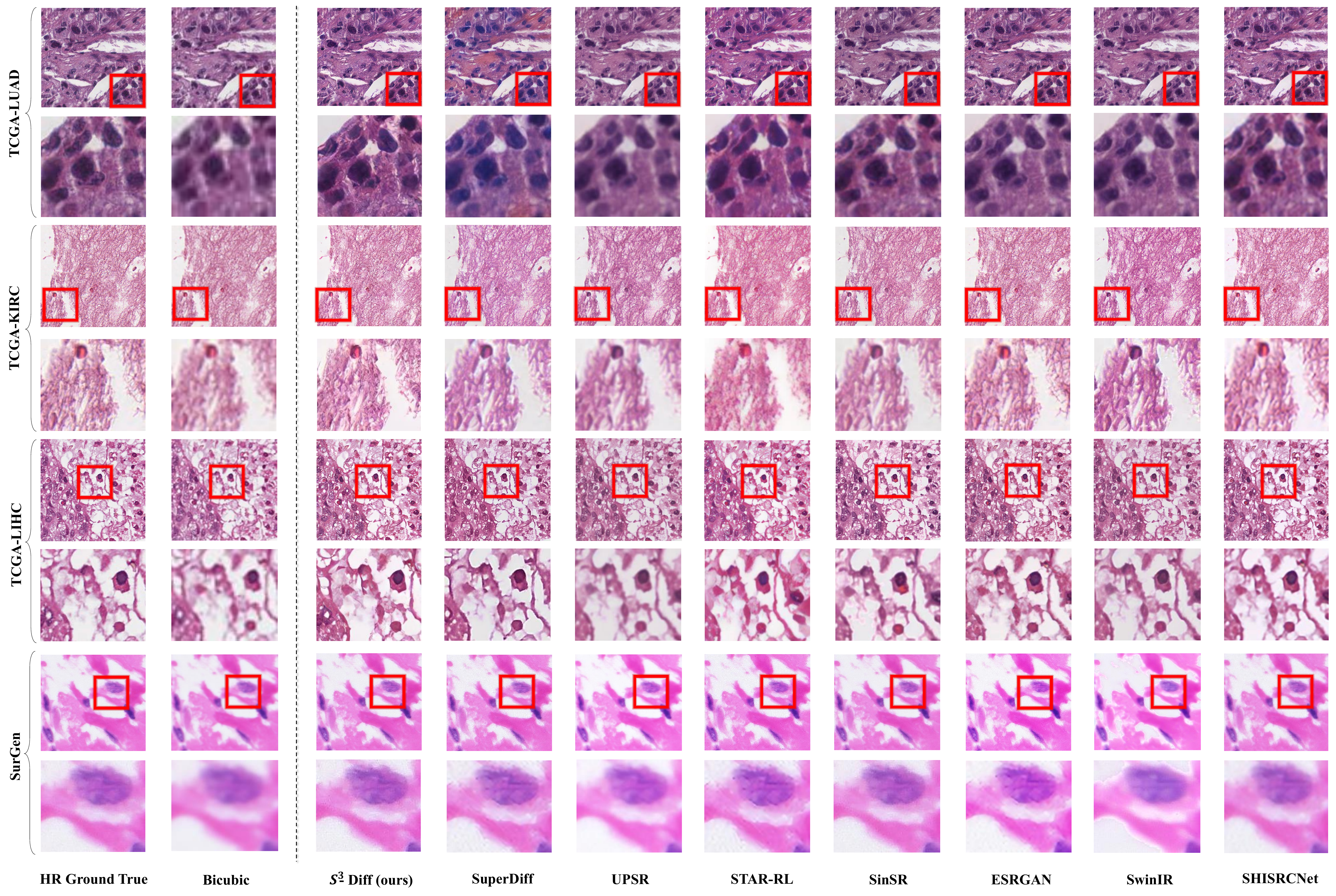}
    \caption{Qualitative Visualizations of SR in pathological images.}
    \label{fig: QV}
\Description{Visual comparison of LR inputs, reconstructed pathology images from competing methods, the S3-Diff output, and the HR reference.}
\end{figure*}

\textbf{Quantitative Analysis of SR}: 
Across the TCGA datasets, S$^3$-Diff consistently achieved the best LPIPS and ST-LPIPS results, as reported in Table~\ref{tab:pre}. The comparison begins with the deterministic end-to-end reconstruction methods. Relative to SwinIR, the strongest deterministic perceptual baseline, S$^3$-Diff reduced LPIPS by 31.89\%, 10.07\%, and 15.89\%, and reduced ST-LPIPS by 33.29\%, 10.76\%, and 16.17\% on TCGA-LUAD, TCGA-KIRC, and TCGA-LIHC, respectively. Although deterministic mappings and the non-learning Bicubic baseline retained competitive pixel-level fidelity, their perceptual errors remained substantially higher. Figure~\ref{fig: QV} provides consistent visual evidence, showing that their conservative reconstruction behavior tends to produce overly smooth images, blurred nuclear boundaries, and attenuated stromal textures.

In contrast, generative frameworks recover richer high-frequency content but remain susceptible to unstable textures and anatomically implausible artifacts. The strongest generative competitors were STAR-RL on TCGA-LUAD, SuperDiff on TCGA-KIRC, and ESRGAN on TCGA-LIHC. Relative to these methods, S$^3$-Diff reduced LPIPS by 23.49\%, 5.80\%, and 20.58\%, respectively, with corresponding ST-LPIPS reductions of 23.63\%, 6.23\%, and 20.84\%. These improvements demonstrate that the perceptual advantage of S$^3$-Diff cannot be attributed solely to the incorporation of generative priors. The combination of SSA-based structural supervision and SSFT-based semantic control establishes a more effective balance between detail recovery and perceptual fidelity. Grad-L1 provides supplementary boundary-sensitive evidence, with S$^3$-Diff ranking first on TCGA-LUAD and TCGA-LIHC and second on TCGA-KIRC. Figure~\ref{fig: QV} further demonstrates that S$^3$-Diff preserves nuclear morphology and stromal textures while limiting both excessive smoothing and conspicuous generative artifacts.

External generalization was further evaluated on the independent SurGen dataset. After being trained exclusively on the TCGA cohorts, S$^3$-Diff was directly transferred to SurGen without any fine-tuning, while this dataset remained entirely excluded from model training and checkpoint selection. As reported in Table~\ref{tab:pre}, S$^3$-Diff achieved SOTA ST-LPIPS and Grad-L1 scores of 0.1711 and 0.0032, respectively, while its LPIPS remained within 1.28\% of the best result obtained by UPSR. Collectively, these results demonstrate that S$^3$-Diff preserves a consistently competitive perceptual profile under direct cross-cohort transfer beyond the TCGA cohorts.

\begin{figure}[!t]
    \centering      
    \includegraphics[width=\columnwidth]{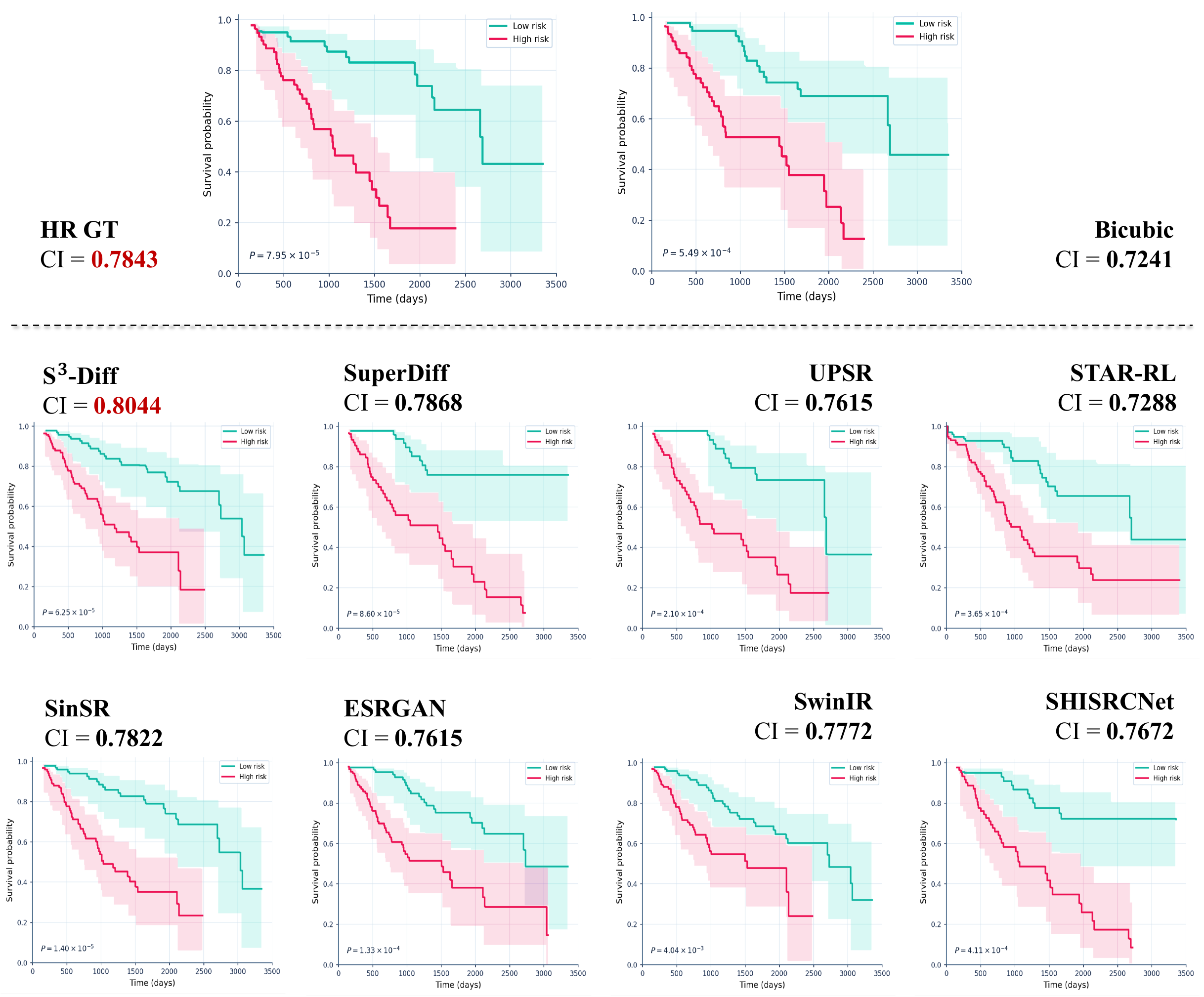}
    \caption{Kaplan-Meier curves comparing survival outcomes from models trained using distinct SR data.}
    \label{fig: KM}
\Description{Kaplan-Meier survival curves comparing risk stratification obtained from models trained on different SR data.}
\end{figure}

\textbf{Downstream Survival Utility}: 
A further evaluation was conducted to determine whether the reconstructed SR images retained prognostically relevant semantic information. All samples were drawn exclusively from the held-out validation cohort and isolated from those used to train the SR models, preventing prognostic information leakage. For each SR method, a generic ABMIL model was independently trained from scratch using only the corresponding reconstructed images. As shown in Figure~\ref{fig: KM}, S$^3$-Diff achieved the most accurate risk ranking, with the highest CI of $0.8044$, substantially exceeding Bicubic interpolation at $0.7241$ and remaining competitive with the HR reference at $0.7843$. The Kaplan--Meier curves further show the clearest separation between the low- and high-risk groups, accompanied by the smallest log-rank p-value of $p=6.25\times10^{-5}$. These results indicate that S$^3$-Diff restores fine-grained structural details in pathological Images while effectively preserving survival-relevant pathological semantics.

\subsection{Ablation and Interpretability Study}

\begin{figure}[!t]
    \centering      
    \includegraphics[width=\linewidth]{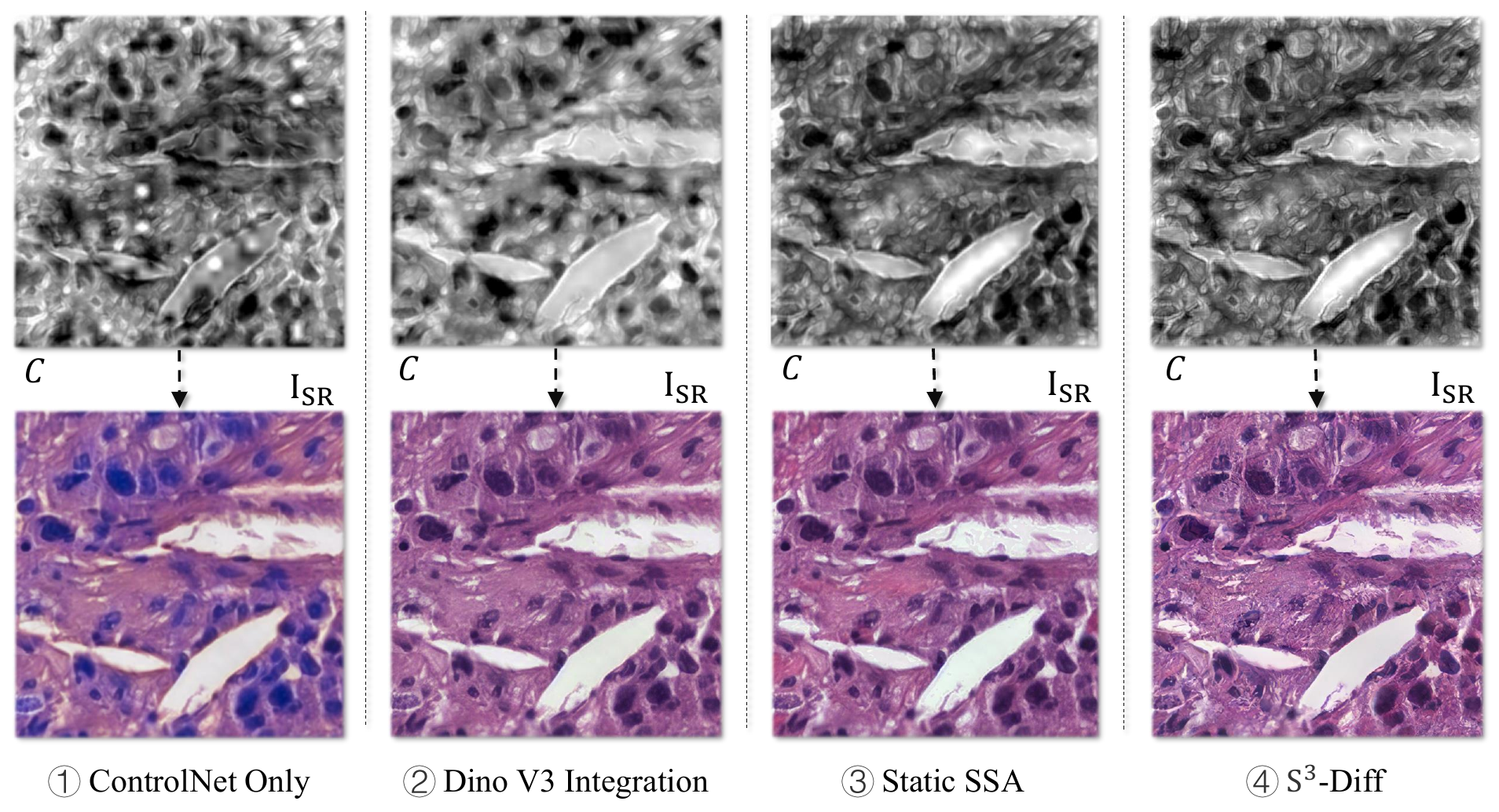}
    \caption{Interpretability analysis for different framework settings.}
    \label{fig: AS}
\Description{Interpretability visualizations comparing control signals and reconstructed tissue regions across the four ablation settings.}
\end{figure}

Table~\ref{tab: AS} summarizes the ablation results of different framework configurations on the TCGA-LUAD dataset. Setting~\textcircled{1} corresponds to the baseline, where only ControlNet is trained within the diffusion model framework. In contrast, setting~\textcircled{2} introduces a fine-tuned DINOv3 model for collaborative training. The results show that this collaborative training substantially improves perceptual metrics such as LPIPS. In addition, the normalized composite control flow $C$ and the corresponding SR results in Figure~\ref{fig: AS} provide visual evidence for the effectiveness of the SSFT strategy. Specifically, synchronized semantic guidance enables more structure-aware restoration and higher-fidelity generation. Furthermore, settings~\textcircled{3} and~\textcircled{4} introduce training-only structural supervision. Here, $A_{str}$ weights latent denoising, while $S_{str}$ weights image fidelity and supervises semantic energy, directing optimization toward critical cellular structures. Compared with setting~\textcircled{3}, setting~\textcircled{4} additionally integrates a prognostic risk coefficient $R$. As shown in Figure~\ref{fig: AS}, this mechanism further improves the precision of morphological restoration by prioritizing clinically and pathologically significant regions.

\begin{table}[!t]
\centering
\renewcommand{\arraystretch}{1.3}
\caption{Performance comparison of different training strategies within the S$^3$-Diff framework on the TCGA-LUAD dataset. The highlighting in this table is the same as in Table \ref{tab:pre}. The evaluated configurations include: \textcircled{1}:  training only the ControlNet for guided generation; \textcircled{2}: fine-tuning DINOv3 without the introduction of $S_str$; \textcircled{3}: incorporating $C$ without the prognostic risk coefficient $R$; and \textcircled{4}: the complete S$^3$-Diff training pipeline.}
\resizebox{\linewidth}{!}{
\begin{tabular}{cccc|ccccc}
\toprule[0.8mm]
\multicolumn{4}{c|}{Framework Setting} & \multicolumn{5}{c}{Metrics}                                                            \\ \cline{5-9} 
\textcircled{1}   & \textcircled{2}   & \textcircled{3}   & \textcircled{4}   & LPIPS\ensuremath{\downarrow}           & ST-LPIPS\ensuremath{\downarrow}        & Grad-L1\ensuremath{\downarrow}         & PSNR\ensuremath{\uparrow}           & SSIM\ensuremath{\uparrow}            \\ \midrule[0.5mm]
\checkmark      &        &       &              & 0.2824          & 0.3014          & 0.0308          & 21.38          & 0.5968          \\
\checkmark      & \checkmark      &       &              & 0.2238 & 0.2243 & 0.0152          & 25.64 & 0.6856 \\
\checkmark     & \checkmark      & \checkmark     &              & {\ul 0.1724}    & {\ul 0.1731}    & {\ul 0.0108}    & {\ul 27.82}    & {\ul 0.7662}    \\
\cellcolor[HTML]{EFEFEF}\checkmark     &\cellcolor[HTML]{EFEFEF}\checkmark      & \cellcolor[HTML]{EFEFEF}\checkmark    & \cellcolor[HTML]{EFEFEF}\checkmark           & \cellcolor[HTML]{EFEFEF}\textbf{0.1651} & \cellcolor[HTML]{EFEFEF}\textbf{0.1645} & \cellcolor[HTML]{EFEFEF}\textbf{0.0073} & \cellcolor[HTML]{EFEFEF}\textbf{29.18} & \cellcolor[HTML]{EFEFEF}\textbf{0.8199} \\ \bottomrule[0.8mm]
\end{tabular}
}
\label{tab: AS}
\end{table}

\section{Limitations and Ethical Considerations}
This study evaluates a limited set of cancer cohorts under a fixed $4\times$ degradation setting, so generalization to untested acquisition conditions remains uncertain. In addition, S$^3$-Diff is intended for research and cannot replace pathologist review or support clinical decisions without prospective multicenter validation. All data use must follow applicable governance requirements. Potential biases should be assessed before deployment, and generated SR images must be identified as computational reconstructions.

\section{Generative AI Usage}
Generative AI assisted language refinement and paper template formatting. It did not generate experimental data or determine reported results and scientific conclusions. The authors verified the manuscript and remain responsible for its accuracy.

\section{Conclusion}
In this work, we propose S$^3$-Diff, a diffusion framework that jointly leverages structural and semantic information for high-fidelity pathology image SR. To reduce morphological distortion and semantic inconsistency, we design SSA to combine prognosis-aware SAM tissue support with LR-HR gradient discrepancies and construct specimen-specific structural references for training-time supervision. In addition, SSFT adapts DINOv3 representations under this supervision and combines semantic energy with LR-derived edge and grayscale cues to guide latent denoising, thereby suppressing artifacts and preserving morphological consistency. Experiments on multiple pathological image datasets demonstrate that S$^3$-Diff achieves stronger perceptual quality and structural fidelity than the compared methods. Beyond improving visual clarity, S$^3$-Diff retains semantic information for downstream analysis, thereby helping bridge computational enhancement and morphologically faithful image reconstruction. In future studies, S$^3$-Diff will be evaluated across multicenter cohorts under real-world acquisition conditions to further assess its generalizability.

\appendix
\section{Implementation and Reproducibility}

\subsection{Optimization Details}

The trainable components specified in the main paper are optimized with AdamW across 60,000 micro-batch iterations. Each iteration processes one paired low-resolution (LR) and high-resolution (HR) image, and gradients accumulate over two iterations, yielding an effective batch size of two and 30,000 optimizer updates. The ControlNet and Energy Head parameter group uses a learning rate of $\eta_c=3\times10^{-6}$, whereas the adapted DINOv3 parameter group uses $\eta_v=6\times10^{-7}$. AdamW uses momentum coefficients of $(0.9,0.999)$, numerical stability constant $10^{-8}$, and weight decay $10^{-2}$. Training uses FP16 precision and random seed 2025. Validation and checkpoint archiving occur every 1,000 micro-batch iterations. The reported checkpoint is selected by the lowest LPIPS on the case-level TCGA validation split, without accessing the TCGA test split or the SurGen cohort. Before vision foundation model (VFM) encoding, the bicubic-upsampled LR reference $I_{up}$ is resized to $256\times256$. Its RGB channels are normalized using ImageNet mean $(0.485,0.456,0.406)$ and standard deviation $(0.229,0.224,0.225)$. Both training and inference use the encoded empty prompt as fixed text conditioning. The Energy Head consists of a $1\times1$ projection from 768 to 64 channels, a $3\times3$ convolution with 64 channels, and a final $1\times1$ projection to one channel. GELU activations follow the first two layers, while sigmoid activation and per-image min--max normalization produce the semantic energy map.

\begin{algorithm}[t]
\caption{Training and inference procedures of S$^3$-Diff}
\begin{algorithmic}[1]
\Require Paired LR and HR images $(I_{lr},I_{hr})$
\Require Precomputed structural reference map $A_{str}$
\Require VAE encoder $\Phi_{enc}$, decoder $\Phi_{dec}$, and U-Net $\Phi_{un}$
\Require ControlNet $\Phi_{ctl}$ and trainable SSFT components

\Statex \textbf{Training}
\For{each micro-batch iteration}
    \State Sample $(I_{lr},I_{hr},A_{str})$
    \State $I_{up}\gets[I_{lr}]_{bic}$
    \State Construct $C$ using Equation~(5) of the main paper
    \State $Z_0\gets\Phi_{enc}(I_{hr})$
    \State Sample diffusion timestep $t$ and $\epsilon\sim\mathcal{N}(0,\mathbf{I})$
    \State Construct $Z_t$ using Equation~(1) of the main paper
    \State Obtain multi-scale residuals $\{B_j\}\gets\Phi_{ctl}(Z_t,t,C)$
    \State Predict conditional noise $\widehat{\epsilon}_{C}\gets\Phi_{un}(Z_t,t,\{B_j\})$
    \State Construct $S_{str}$ using Equation~(6) of the main paper
    \State Recover $\widehat{Z}_0$ and $I_{rec}$ using Equation~(7) of the main paper
    \State Compute $\mathcal{L}_{tot}$ using Equation~(8) of the main paper
    \State Accumulate gradients from $\mathcal{L}_{tot}$
\If{two iterations have been accumulated}
    \State Update $\Phi_{ctl}$ and the trainable SSFT parameters
\EndIf
\EndFor

\Statex \textbf{Inference with fixed parameters}
\State $I_{up}\gets[I_{lr}]_{bic}$
\State Construct $C$ using Equation~(5) of the main paper
\State Sample $\epsilon\sim\mathcal{N}(0,\mathbf{I})$
\State Initialize the working latent $Z\gets Z_{t_s}$ using Equation~(A.1)
\For{$t\in\{t_k\}_{k=K-K_i+1}^{K}$}
    \State Obtain multi-scale residuals $\{B_j\}\gets\Phi_{ctl}(Z,t,C)$
    \State Predict conditional noise $\widehat{\epsilon}_{C}\gets\Phi_{un}(Z,t,\{B_j\})$
    \State Update $Z$ through one DDPM reverse step using $\widehat{\epsilon}_{C}$
\EndFor
\State $\widetilde{I}_{sr}\gets\Phi_{dec}(Z)$
\State Obtain $I_{sr}$ using Equation~(A.2)
\State \Return $I_{sr}$
\end{algorithmic}
\label{al: 1}
\end{algorithm}

\subsection{Details of LR-Anchored Inference}

LR-anchored inference initializes reverse diffusion by jointly encoding and perturbing the bicubic reference image. Let $I_{lr}$ denote the LR input and $I_{up}$ its bicubic interpolation to the target HR dimensions. Let $\{t_k\}_{k=1}^{K}$ denote the descending DDPM timestep sequence, where $K$ is the number of scheduled sampling steps. The initialization strength $\xi_i\in(0,1]$ determines the retained step count $K_i$, the initial timestep $t_s$, and the initial noisy latent:
\begin{equation}
\begin{gathered}
K_i
=
\left[
[\xi_iK]_{rnd}
\right]_{[1,K]},
\qquad
t_s
=
t_{K-K_i+1},
\\
Z_{t_s}
=
\sqrt{\bar{\alpha}_{t_s}}\,
\Phi_{enc}(I_{up})
+
\sqrt{1-\bar{\alpha}_{t_s}}\epsilon.
\end{gathered}
\tag{A.1}
\end{equation}
Here, $\Phi_{enc}(\cdot)$ denotes the VAE encoder, and $\Phi_{enc}(I_{up})$ is the encoded LR-reference latent. The operator $[\cdot]_{rnd}$ denotes nearest-integer rounding, while $[\cdot]_{[1,K]}$ bounds the retained step count to the valid range. The coefficient $\bar{\alpha}_{t_s}$ denotes cumulative noise retention at $t_s$, and $\epsilon\sim\mathcal{N}(0,\mathbf{I})$ is Gaussian noise with identity covariance $\mathbf{I}$. By default, $K=30$ and $\xi_i=0.20$, giving $K_i=6$ and $t_s=166$. The retained reverse timesteps are $\{166,133,100,67,34,1\}$.

The LR-derived composite control flow $C$, composed of semantic-energy, edge, and grayscale cues in Equation~(5) of the main paper, conditions every retained reverse step. After reverse diffusion produces the clean latent $Z_0$, the VAE decoder $\Phi_{dec}(\cdot)$ generates the preliminary image $\widetilde{I}_{sr}$. The two-stage LR-anchored reconstruction in Equation~(9) of the main paper is implemented as
\begin{equation}
\begin{gathered}
H_1
=
\widetilde{I}_{sr}
-
[\widetilde{I}_{sr}]_{lp}^{(5)},
\\
I_{sr}^{(1)}
=
\left[
I_{up}
+
2.10H_1
\right]_{[0,1]},
\\
H_2
=
\left[
I_{sr}^{(1)}
-
[I_{sr}^{(1)}]_{lp}^{(7)}
\right]_{[-0.12,0.12]},
\\
I_{sr}
=
\left[
I_{up}
+
1.15H_2
\right]_{[0,1]}.
\end{gathered}
\tag{A.2}
\end{equation}
Here, $[\cdot]_{lp}^{(5)}$ and $[\cdot]_{lp}^{(7)}$ denote reflection-padded $5\times5$ and $7\times7$ mean filters, respectively, while an interval subscript denotes element-wise clipping. The variables $H_1$ and $H_2$ represent the first- and second-stage high-frequency residuals, respectively. The first stage transfers diffusion-derived detail to the LR reference using a default gain of $2.10$. The second stage constrains its residual to $[-0.12,0.12]$ before applying a default gain of $1.15$ and restoring the LR anchor. The reported $I_{sr}$ is obtained directly from this reconstruction, with color correction disabled.

\subsection{Pseudocode Implementation}

Algorithm~\ref{al: 1} summarizes the training and inference procedures. It uses the main-paper equations by their fixed numbers and the locally defined inference Equations~(A.1) and~(A.2).

Here, $[\cdot]_{bic}$ denotes bicubic interpolation, and $B_j$ denotes the residual produced at ControlNet scale $j$. The variable $Z$ represents the working latent during reverse diffusion, while $\widehat{\epsilon}_{C}$ denotes the conditional noise predicted by the frozen U-Net under ControlNet guidance. The control flow $C$ is constructed once for each input and reused across all retained reverse steps.

\subsection{Software and Runtime Environment}

All experiments were implemented in PyTorch and executed on a single NVIDIA RTX PRO 6000 GPU. Stable Diffusion v1.5~\cite{rombach2022high} provides the latent-diffusion backbone, including the frozen VAE,
text encoder, and U-Net. ControlNet is initialized from the pretrained U-Net. DINOv3 ViT-B/16~\cite{simeoni2025dinov3} and the Energy Head
construct the semantic control signal, while SAM
ViT-B~\cite{kirillov2023segment} generates structural reference maps during offline preprocessing. Training updates ControlNet, the Energy Head, the final DINOv3 Transformer block, and all DINOv3 LayerNorm parameters. The exact optimization and LR-anchored inference settings are specified in Sections~A.1 and~A.2. Inference requires only the LR input and does not use SAM, an HR target, prognostic metadata, or the training-time structural reference map $A_{str}$.

\section{Data Preparation and Validation Protocol}

\subsection{TCGA Cohorts}

To evaluate robustness across heterogeneous histopathological patterns, we
curated a multi-organ cohort from The Cancer Genome Atlas
(TCGA)~\cite{weinstein2013cancer}. The cohort comprises 150 H\&E-stained
whole-slide images (WSIs), with 50 WSIs randomly selected from each of
TCGA-LUAD, TCGA-KIRC, and TCGA-LIHC. We partitioned the data at the patient
level into training, validation, and test sets in a 7:1:2 ratio. Partitioning
was performed before patch extraction to ensure that no patient contributed
patches to more than one subset.

\begin{figure*}[t]
    \centering
    \includegraphics[width=\textwidth,trim=0 90mm 0 0,clip]{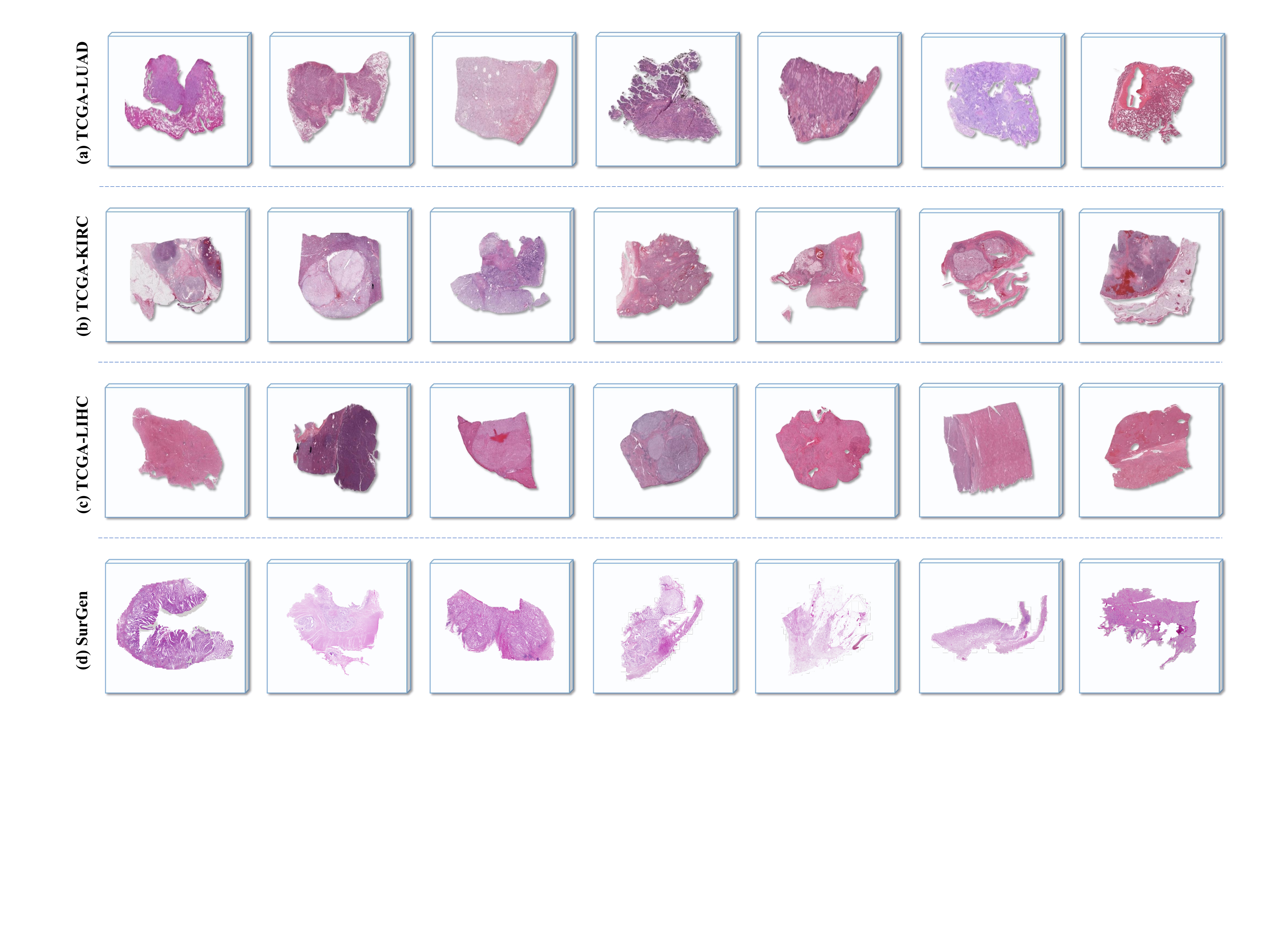}
    \caption{Representative whole-slide images from the TCGA cohorts. Rows show examples from (a) TCGA-LUAD, (b) TCGA-KIRC, (c) TCGA-LIHC, and (d) SurGen,
    illustrating inter-slide variation in tissue morphology and staining.}
    \label{fig:tcga-overview}
\end{figure*}

The cohorts span distinct tissue morphologies: heterogeneous glandular growth
and nuclear atypia in LUAD; clear cytoplasm and abundant microvasculature in
KIRC; and disrupted lobular architecture with trabecular or pseudoglandular
growth in LIHC. Representative WSIs from the three cohorts are shown in
Figure~\ref{fig:tcga-overview}.

\subsection{Independent External Validation Cohort}

We further employed the publicly available SurGen dataset (BioStudies accession S-BIAD1285) as an independent external validation cohort~\cite{myles2025surgen}. SurGen is a colorectal
cancer histopathology resource comprising H\&E-stained WSIs with matched survival information. The released protocol evaluates all available matched slides in the prepared cohort and retains at most
300 paired patches per slide using deterministic stable-hash sampling with seed 2025. The TCGA-trained checkpoint remains frozen throughout this evaluation. SurGen contributes neither to model training, checkpoint selection, SAM-based structural-reference generation, nor hyperparameter optimization. During external inference, only LR images enter S$^3$-Diff, while HR images serve exclusively as evaluation references.

The selected SurGen slides were processed using the same tissue-detection,
patch-extraction, and $4\times$ degradation pipeline as the TCGA slides.
SurGen was not used for model training, validation-based model selection,
checkpoint selection, or hyperparameter optimization. It served exclusively
as an external test cohort for assessing reconstruction quality and
cross-dataset generalization.

\subsection{Patch Extraction and Degradation}

We used the CLAM preprocessing pipeline~\cite{lu2021data} to identify tissue
regions and extracted 34,495 valid high-resolution (HR) patches from the TCGA
WSIs. Each HR patch contained $512\times512$ pixels and corresponded to a
field of view at $40\times$ optical magnification. Each patch was then bicubically downsampled by a factor of four to generate a paired
$128\times128$ low-resolution (LR) input. The same spatial pairing and image-size checks were applied to the available matched SurGen slides.

\section{Evaluation Metrics}

We evaluated both super-resolution image quality and downstream clinical
utility. Lower LPIPS, ST-LPIPS, and Grad-L1 values indicate better
performance, whereas higher PSNR, SSIM, and concordance index values indicate
better performance.

\subsection{Signal and Structural Fidelity}

For images with maximum attainable intensity $L$, peak signal-to-noise ratio
(PSNR)~\cite{wang2004image} is defined as
\begin{equation}
    \operatorname{PSNR}(x,y)
    =
    10\log_{10}
    \left(
        \frac{L^2}{\operatorname{MSE}(x,y)}
    \right).
\end{equation}
The structural similarity index (SSIM) compares local luminance, contrast, and
structure:
\begin{equation}
    \operatorname{SSIM}(x,y)
    =
    \frac{
        (2\mu_x\mu_y+c_1)(2\sigma_{xy}+c_2)
    }{
        (\mu_x^2+\mu_y^2+c_1)
        (\sigma_x^2+\sigma_y^2+c_2)
    }.
\end{equation}
We treated PSNR and SSIM as complementary measures of pixel-level fidelity,
because high pixel-wise similarity does not necessarily imply faithful
reconstruction of fine pathological morphology.

\subsection{Perceptual and Morphological Consistency}

Learned Perceptual Image Patch Similarity
(LPIPS)~\cite{zhang2018perceptual} measures perceptual distance in the feature
space of a pretrained network:
\begin{equation}
    d(x,x_0)
    =
    \sum_l
    \frac{1}{H_lW_l}
    \sum_{h,w}
    \left\|
        w_l\odot
        \left(
            \hat{y}_{hw}^{\,l}
            -
            \hat{y}_{0,hw}^{\,l}
        \right)
    \right\|_2^2.
\end{equation}
We additionally used ST-LPIPS~\cite{ghildyal2022shift}, a shift-tolerant
perceptual measure that reduces sensitivity to small spatial translations
between the reconstruction and its reference.

Grad-L1 compares the aggregate first-order horizontal and vertical RGB differences of a reconstruction and its paired HR target. For an image $x$, we compute \begin{equation} g(x) = \operatorname{mean}(|x_{:,:, :,2:}-x_{:,:, :,:-1}|) + \operatorname{mean}(|x_{:,:,2:,:}-x_{:,:,:-1,:}|), \end{equation} and report \begin{equation} \operatorname{Grad	ext{-}L1}(x,y) = |g(x)-g(y)|. \end{equation}

\subsection{Downstream Survival Utility}

We used an attention-based multiple-instance learning
model~\cite{ITW:2018} to assess whether the reconstructed images retained
prognostically relevant pathological information. The concordance
index~\cite{wang2019machine} quantifies agreement between predicted risk
rankings and observed survival outcomes. For survival time $T_i$, predicted
risk $\eta_i$, and event indicator $\delta_i$, the concordance index is
computed as
\begin{equation}
    \operatorname{CI}
    =
    \frac{
        \sum_{i,j}
        \mathbb{I}(T_i<T_j)
        \mathbb{I}(\eta_i>\eta_j)
        \delta_i
    }{
        \sum_{i,j}
        \mathbb{I}(T_i<T_j)\delta_i
    }.
\end{equation}
Kaplan--Meier estimators were used to visualize the survival distributions of
the predicted low- and high-risk groups. Their separation was assessed using
a log-rank test.

\end{document}